\documentclass[10pt,twocolumn,letterpaper]{article}

\usepackage{wacv}

\usepackage[accsupp]{axessibility}
\usepackage{graphicx}
\usepackage{amsmath}
\usepackage{amssymb}
\usepackage{booktabs}
\usepackage{graphicx}
\usepackage{booktabs}
\usepackage{makecell}
\usepackage{comment}
\usepackage{algorithm}
\usepackage{algorithmic}
\usepackage{amsfonts}       
\usepackage{amssymb}  
\usepackage{array}
\usepackage{xcolor}
\usepackage{tikz}
\usetikzlibrary{arrows.meta, positioning}
\usetikzlibrary{fit}
\usepackage{graphicx}
\usepackage{pifont}
\newcommand{\cmark}{\ding{51}}%
\newcommand{\xmark}{\ding{55}}%
\usepackage{arydshln}   
\usepackage[pagebackref,breaklinks,colorlinks]{hyperref}

\usepackage[capitalize]{cleveref}
\crefname{section}{Sec.}{Secs.}
\Crefname{section}{Section}{Sections}
\Crefname{table}{Table}{Tables}
\crefname{table}{Tab.}{Tabs.}

\begin{document}

\title{Food Image Generation on Multi-Noun Categories}

\author{Xinyue Pan\\
Purdue University\\
{\tt\small pan161@purdue.edu}
\and
Yuhao Chen $^{\dagger}$\\
University of Waterloo\\
{\tt\small yuhao.chen1@uwaterloo.ca}
\and
Jiangpeng He\\
Purdue University\\
{\tt\small he416@purdue.edu}
\and
Fengqing Zhu\\
Purdue University\\
{\tt\small zhu0@purdue.edu}
}
\maketitle

\begin{abstract}
Generating realistic food images for categories with multiple nouns is surprisingly challenging. For instance, the prompt ``egg noodle" may result in images that incorrectly contain both eggs and noodles as separate entities. Multi-noun food categories are common in real-world datasets and account for a large portion of entries in benchmarks such as UEC-256. These compound names often cause generative models to misinterpret the semantics, producing unintended ingredients or objects. This is due to insufficient multi-noun category related knowledge in the text encoder and misinterpretation of multi-noun relationships, leading to incorrect spatial layouts. To overcome these challenges, 
we propose \textbf{FoCULR} (\textbf{Fo}od \textbf{C}ategory \textbf{U}nderstanding and \textbf{L}ayout \textbf{R}efinement) which incorporates food domain knowledge and introduces core concepts early in the generation process.
Experimental results demonstrate that the integration of these techniques improves image generation performance in the food domain. 

\end{abstract}

\vspace{-3mm}
\section{Introduction}  

\begin{figure}[t]
    \centering
    \includegraphics[width=0.95\linewidth]{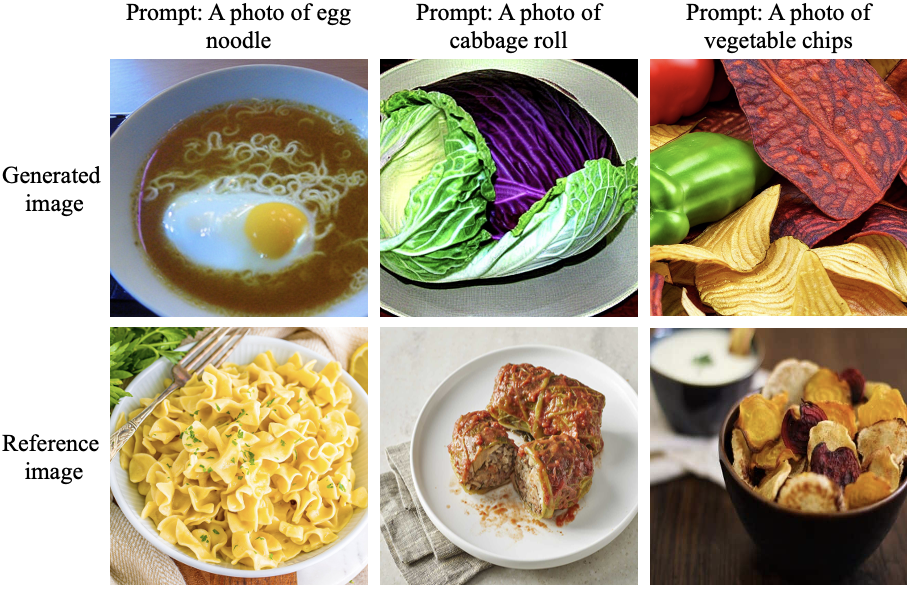}
    \caption{Example multi-noun food images generated by stable diffusion v1-4 and corresponding reference images on selected categories.}
    \label{fig:empirical}
\vspace{-3mm}
\end{figure}

Text-to-image generation models \cite{nichol2021glide,Rombach_2022_CVPR,saharia2022,ramesh2022hierarchical, ho2022classifier,zhong2023adapter} have achieved remarkable success for single-object prompts as input to generate images, but they routinely fail when asked to generate multi-noun objects, e.g., “egg noodle,” “cabbage roll,” “vegetable chips”, as shown in Figure \ref{fig:empirical}. In these cases, the model often renders each noun as separate objects, splitting “egg” and “noodle” rather than producing a single, coherent food item.  

From the application perspective, multi-noun foods are ubiquitous in menus, recipes, and dietary records. Additionally, in a typical dataset such as the UEC-256\cite{kawano14c}, multi-noun food categories make up about 50\% of all classes in the dataset. Accurately generating such compound food is critical for downstream tasks such as virtual menu previews, personalized nutrition advice, and augmented reality recipe assistants, all of which require generating food objects with multi-noun categories\cite{vir2024archef}. Thus, solving the multi-noun food image generation problem has both technical and practical importance.

Previous work in image generation~\cite{karras2021, Rombach_2022_CVPR, ruiz2023, Zhang_2023_ICCV, zheng2023layoutdiffusion, wang2024instancediffusion} has largely focused on enhancing overall image quality or introducing various forms of conditional control (e.g., style or text guidance) using Stable Diffusion model\cite{Rombach_2022_CVPR}. However, relatively little attention has been given to the challenge of generating coherent images from multi-noun categories. Common domain adaptation approaches involve fine-tuning only the model’s image generation network, leaving the text encoder frozen. Yet, this strategy often leads to redundant layouts during generation process, as each noun within the prompt tends to be treated as an independent object. Our analysis identifies two primary factors leading to this issue. \textbf{First}, the frozen text encoder and image generation model lack knowledge related to multi-noun categories (see details in Section A of the supplementary material). 
\textbf{Second}, the text encoder inherently struggles to unify multiple nouns semantically, leading to incorrect attention weights on individual tokens and consequently erroneous layout initialization.

To address these challenges, we propose \textbf{FoCULR} (\textbf{Fo}od \textbf{C}ategory \textbf{U}nderstanding and \textbf{L}ayout \textbf{R}efinement), which integrates two complementary components including (i) \textbf{Food Domain Adaptation via Local Alignment (FDALA)}, and (ii) \textbf{Core-Focused Image Generation (CFIG)}. Specifically, \textbf{FDALA} fine-tunes the attention layers of both the text encoder and image generation network within the Stable diffusion model to learn domain-specific food knowledge, which includes many multi-noun categories, thus improving the model’s understanding of multi-noun food prompts.
\textbf{CFIG} is applied during inference, where negative prompts (i.e., instructions to exclude specific objects) help focus on the main food structure in the early timesteps, before refining details. Additionally, the \textbf{CFIG} alone does not require additional computational resource. 

Our contributions can be summarized as follows.
\begin{itemize}
    \item To our knowledge, this is the first work to address multi-noun category image generation in the food domain.
    \item We propose \textbf{Food Domain Adaptation via Local Alignment (FDALA)} to enable the text encoder in the Stable diffusion model to learn domain-specific knowledge related to food.
    \item We introduce \textbf{Core-Focused Image Generation (CFIG)}, an approach that applies negative prompts only during early timesteps of the inference phase to guide the model to first generate the main structure (shape) of the food object, followed by the refinement of fine-grained details in later timesteps.
\end{itemize}

\vspace{-1mm}

\section{Related Work}

\subsection{Food Image Generation}
There is an extensive literature sought to improve alignment between visual content and textual descriptions in food image generation. 
Han \textit{et al.} demonstrated that diffusion models outperform GAN (Generative Adversarial Network) in capturing the intra‑class diversity inherent in varied cooking recipes, thereby improving overall image quality \cite{han2023}. Yamamoto \textit{et al.} leveraged CLIP \cite{radford2021learning} embeddings to enforce semantic consistency in food image editing tasks \cite{Yamamoto2022}, while Li \textit{et al.}’s ChefFusion framework combined a Transformer‑based language model with diffusion for joint recipe generation and image synthesis \cite{li2024cheffusion}. Ma \textit{et al.} introduced MLA‑Diff, which enhances CLIP‑based \cite{radford2021learning} ingredient–image embeddings through an attention‑fusion module to boost realism \cite{ma2024}. More recently, Yu \textit{et al.} proposed CW‑Food, a diffusion pipeline that decouples shared intra‑class features and private instance details via Transformer fusion and LoRA fine‑tuning, enabling diverse image synthesis from food names alone \cite{yu2025}. However, none of these approaches directly address the multi‑noun category problem
, motivating the development of our FoCULR framework.

\subsection{Attribute Binding in Image Generation}
\label{sec:rel_attr_bind}
Attribute binding refers to associating attributes (e.g., color, size) with corresponding objects in generated images \cite{wu2022adma, feng2023trainingfree, rassin2024linguistic, clark2023neurips,trusca2024object,taghipour2024box}. For example, given the prompt ``a red car and a white house," an attribute binding error would result in swapping the colors, generating a white car and a red house. 
This problem is close to the multi-noun category problem in food image generation. In both cases, a model must correctly understand the relationship between words. However, while attribute binding typically involves modifiers (e.g., ``red car" and ``white house"), the multi-noun category problem involves compound nouns, where multiple words (e.g., ``egg sandwich") must be interpreted as a single entity rather than separate objects.
Among the previously proposed solutions, two training-free methods are particularly relevant, as they are applied during the denoising steps of the Stable diffusion model, close to our proposed \textbf{CFIG}. 
Feng \textit{et al.} \cite{feng2023trainingfree} introduced structured guidance by parsing text prompts into paired entities 
to ensure correct associations during image generation. Similarly, Royi \textit{et al.} \cite{rassin2024linguistic} proposed a syntactic parsing method with a custom loss function to align related words and prevent incorrect attribute-object pairings.
Although these methods address the attribute binding problem, they do not tackle issues arising from multi-noun category misinterpretation. Instead of focusing on attributes like color or size, the multi-noun category problem involves distinguishing between semantically overlapping components. 
This distinction motivates the need for a more domain-specific approach, such as our proposed modules, which address the generation of redundant food objects by improving text encoder fine-tuning and layout initialization.

\vspace{-1mm}

\section{Methods}
\begin{figure*}[t]
    \centering
    \includegraphics[width=1\linewidth]{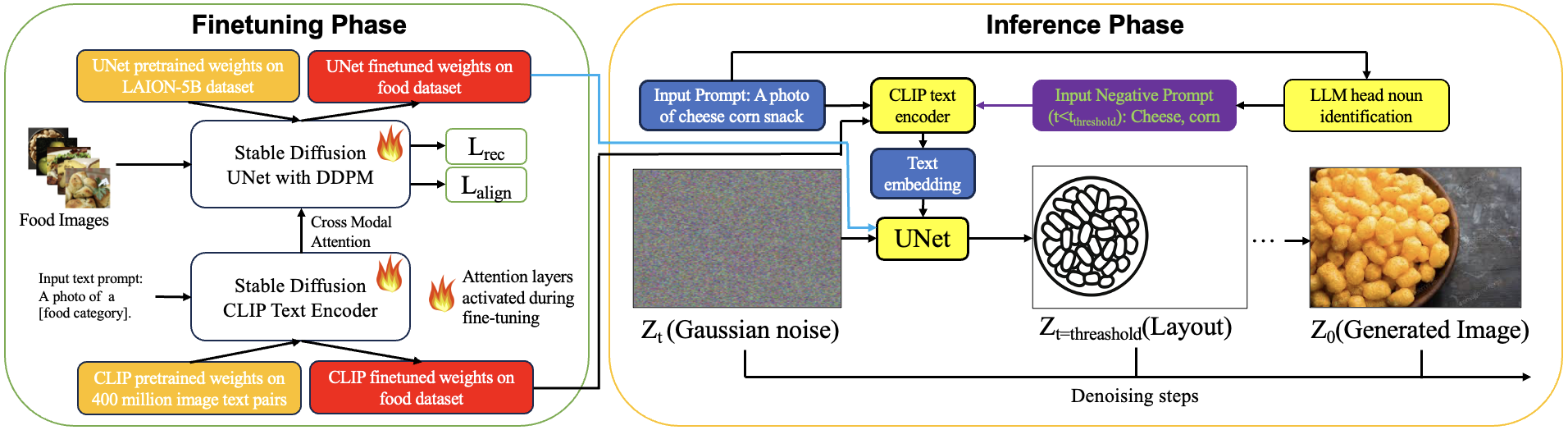}

    \caption{Overview of our method \textbf{FoCULR}: During fine-tuning phase, the pretrained weights on LAION-5B dataset for stable diffusion UNet model and pretrained weights on 400 million image-text pairs for CLIP model \cite{radford2021learning} are loaded for finetuning. The loss function is a combination of image-concept alignment loss $L_{align}$ and reconstruction loss $L_{rec}$ that learns food domain knowledge on both the UNet and the text encoder. During inference, negative prompt is activated only when $t<t_{threshold}$ during denoising steps of inference phase. The negative prompt is generated by GPT-4o head noun identification from the input prompt. The denoising steps are scheduled according to the DDPM (Denoising Diffusion Probabilistic Model)\cite{ho2020denoising}.}
    \label{fig:finetune}
\vspace{-3mm}
\end{figure*}


Figure \ref{fig:finetune} provides an overview of our method \textbf{FoCULR}, which comprises two components targeting distinct stages of the generation process: fine-tuning and inference. During fine-tuning, the prompt template “A photo of [category]” is fed to the CLIP text encoder \cite{radford2021learning} (pretrained on 400 million image‑text pairs), and its output embeddings are used via cross‑attention in the UNet (pretrained on LAION‑5B\cite{Schuhmann2022}) to guide the image generation. The model is learned with two loss functions, which are detailed in Section \ref{sec:FDKF}. During the inference phase, to mitigate the occurrence of redundant food objects, we introduce constraints via negative prompts during the initial denoising steps as detailed in Section \ref{sec:CFIG}. 

\subsection{Preliminaries: Latent Diffusion Model}
To contextualize our proposed method, we briefly review the diffusion process in a latent diffusion model \cite{Rombach_2022_CVPR}. To reduce computational costs, the image data is first encoded into a latent space using a variational autoencoder (VAE) \cite{rezende2014stochastic}. Gaussian noise is then added to this latent space, and the reverse process is performed through a UNet \cite{ho2020denoising} to denoise and restore the image. Text inputs are processed using a CLIP text encoder \cite{radford2021learning}, which generates text embeddings to condition and guide the image generation process. During training, the model minimizes the reconstruction loss defined as the expectation between the predicted noise and the added noise at each timestep. More formally, the objective function is defined as:
\vspace{-1mm}
\begin{equation}
    L_{rec} = \mathbb{E}_{z, c, \epsilon, t} \left[ \| \epsilon - \epsilon_{\theta}(z_t, c, t) \|_2^2 \right],
\label{eq:rec_loss}
\vspace{-1mm}
\end{equation}
where \( z \) is the latent representation of the input image obtained via the VAE encoder, \( z_t \) is the noisy latent at timestep \( t \), generated by adding Gaussian noise \( \epsilon \sim \mathcal{N}(0, I) \) to \( z \), \( c \) is the text embedding from the CLIP text encoder \cite{radford2021learning}, \( \epsilon_{\theta}(z_t, c, t) \) is the predicted noise from the UNet model parameterized by \( \theta \), and \( t \) is the diffusion timestep. 

After training, images are generated by sampling Gaussian noise conditioned on a text prompt using a scheduler, such as DDPM (Denoising Diffusion Probabilistic Model)\cite{ho2020denoising}. The scheduler iteratively samples from timestep \( t \) to the previous step \( t-1 \) until \( t=0 \), following this denoising process:
\vspace{-1mm}
\begin{equation}
    z_{t-1} = \frac{1}{\sqrt{\alpha_t}} \left( z_t - \frac{\beta_t}{\sqrt{1 - \bar{\alpha}_t}} \epsilon_{\theta}(z_t, c, t) \right) + \sigma_t \xi,
\vspace{-1mm}
\end{equation}
where \( z_t \) and \( z_{t-1} \) represent the noisy latent at timesteps \( t \) and \( t-1 \), respectively, \( \alpha_t \) is the noise schedule parameter controlling the amount of noise at timestep \( t \), \( \bar{\alpha}_t \) is the cumulative product of \( \alpha_t \) over all timesteps, \( \beta_t \) is the noise variance at timestep \( t \), \( \epsilon_{\theta}(z_t, c, t) \) is the predicted noise from the UNet model, \( \sigma_t \) is the standard deviation of the noise added at timestep \( t \), and \( \xi \) is random Gaussian noise sampled from \( \mathcal{N}(0, I) \).

\subsection{FDALA: Food Domain Adaptation via Local Alignment}
\label{sec:FDKF}
Fine-tuning the text encoder in diffusion models is not common, primarily due to the associated challenges such as high time complexity and high GPU memory requirement\cite{hu2022lora}. However, our empirical studies (see supplementary material Section A) reveal that the pretrained text encoder lacks sufficient food domain knowledge, which contains many multi-noun categories. To address this, we fine-tune both the text encoder and image generation network for better semantic understanding of multi-noun categories. This fine-tuning is implemented on the CLIP text encoder and UNet backbone of Stable Diffusion model. Inspired by LoRA\cite{hu2022lora}, we incorporate lightweight adapters into the attention layers and only fine-tune the attention layers of both networks to preserve the overall model efficiency. 

Food images often exhibit high inter-class similarity, leading to confusions in the generated image between categories such as ``chicken salad" and ``tuna salad", which are visually similar yet semantically distinct. Therefore, it is challenging to use global alignment approaches  \cite{li2024textcraftor}, where the loss function is based on similarity score between global image embedding and global text embedding. To address this challenge, we introduce an image-concept alignment objective inspired by FILIP \cite{yao2022filip} and OVFoodSeg\cite{wu2024ovfoodseg} that ensures semantic coherence between image patches and specific text concepts. Specifically, we align localized image embeddings (e.g., patch-level features) with semantically relevant concept embeddings extracted from text. In practice, we use the CLIP vision encoder to obtain patch-level image features and the CLIP text encoder to obtain embeddings for the category label tokens. This design encourages the model to distinguish subtle visual cues linked to category-defining concepts, thereby improving class discriminability in visually similar food items, i.e. ``chicken salad" and ``tuna salad".

More formally, to find the most relevant image patch related to a concept, we first compute the maximum similarity for each concept \( c \) in all image patches. Let \( M \in \mathbb{R}^{N \times T} \) denote the similarity matrix, where \( M_{c,i} \) is the cosine similarity between the \( i \)-th image patch feature and the \( c \)-th concept text embedding (higher value indicates similar image patch with the concept). The maximum similarity $S_{c}$ can be formulated as following:
\vspace{-1mm}
\begin{equation}
    S_{c} = \max_{i} M_{c,i}
\vspace{-1mm}
\end{equation}
The overall alignment score \( S \) for the image and text is computed as the average of alignment scores across all concepts \( C \):
\vspace{-1mm}
\begin{equation}
S = \frac{1}{|C|} \sum_{c \in C} S_c,
\vspace{-1mm}
\end{equation}
where \( |C| \) is the total number of concepts in the text.
The image-concept alignment loss is then defined as:
\vspace{-1mm}
\begin{equation}
L_{align} = 1 - S
\vspace{-1mm}
\end{equation}
The total loss function \( L_{total} \) used for fine-tuning the UNet and CLIP text encoder \cite{radford2021learning} combines the reconstruction loss (Eq.\ref{eq:rec_loss}) and image-concept alignment loss:
\vspace{-1mm}
\begin{equation}
L_{total} = L_{rec} + L_{align} 
\vspace{-1mm}
\end{equation}
This formulation ensures fine-grained semantic alignments, improving its ability to generate accurate and contextually consistent food images.

\subsection{CFIG: Core-Focused Image Generation}
\label{sec:CFIG}
As described in \cite{yi2024towards}, the denoising process in a latent diffusion model begins with the generation of coarse layout and structural elements. During the early timesteps, the model establishes the spatial structure, while the later timesteps focuses on progressively filling in fine-grained details within this layout. However, in the case of multi-noun categories, this sequential process often lead to the generation of redundant components.
Specifically, during early denoising steps, the model struggles to allocate attention effectively across multiple nouns token. Instead of prioritizing the core object described by the category, the model overemphasizes sub-phrases such as ``egg" in “egg sandwich”.  This misalignment in the early steps leads to inferior spatial initialization and the generation of semantically redundant or extraneous visual elements.

To overcome this challenge, we propose a simple yet effective inference-time strategy tailored to multi-noun categories, called \textbf{Core-Focused Image Generation (CFIG)}. Specifically, we introduce constraints by applying suitable negative prompts during the initial denoising steps ($t < t_{\text{threshold}}$) to suppress non-core components (e.g., ``egg" in ``egg sandwich"), which are used as the input to the negative prompts. During these early steps, our method leverages linguistic insights by identifying the “head noun” of a phrase, as discussed below, and suppresses all other non-head tokens. This compels the model to form a coherent, unified layout based solely on the main noun (e.g., ``sandwich”).  For later steps, only positive prompts are used which allows the model to refine details without suppressing any components, thereby ensuring both semantic fidelity and high-quality visual details. The threshold $t_{\text{threshold}}$ is an exogenous hyperparameter, empirically set to 5 out of a total of 50 inference steps based on ablation studies provided in supplementary material Section C. The adjusted noise prediction at each timestep is thus formulated as:
\vspace{-1mm}
\begin{equation}
\epsilon_{\text{pred}} = 
\begin{cases} 
    \epsilon_{\theta}(z_t, t) 
    + \lambda \big( \epsilon_{\theta}(z_t, c_{\text{pos}}, t) \\ 
    - \epsilon_{\theta}(z_t, c_{\text{neg}}, t) \big), 
    & \text{if } t < t_{\text{threshold}}, \\[8pt]
    \epsilon_{\theta}(z_t, t) 
    + \lambda \big( \epsilon_{\theta}(z_t, c_{\text{pos}}, t)   \big), 
    & \text{if } t \geq t_{\text{threshold}}.
\end{cases}
\end{equation}
where \( \epsilon_{\theta}(z_t, t) \) denotes the \textit{unconditional noise prediction}, 
\( c_{\text{pos}} \) is the \textit{positive prompt}, 
\( c_{\text{neg}} \) is the \textit{negative prompt}, 
and \( \lambda \) is the \textit{guidance scale parameter} (set to \( 7.5 \) in our experiments). This method ensures a focused layout generation in the early stages, followed by refinement of details later.

In English noun phrases, the rightmost noun typically functions as the syntactic ``head," defining the core object and its general shape. For example, in the phrase ``corn chips," the noun ``chips" serves as the head—denoting the primary flat, thin, and crispy shape—held form—while ``corn" acts as a modifier contributing surface detail. We leverage this linguistic insight via a prompt given to a Large Language Model (LLM): ``Given the food item [category], what is the head element that defines its overall shape?" Then, the ``non-head" part is considered as non-core component of the compound nouns and is served as the input to the negative prompt. In this paper, GPT-4o\cite{hurst2024gpt} is used due to its extensive real‑world, multi-modal pretraining that ensures broad generalizability across food categories and datasets. 

Note that the CFIG module can be independently applied without incorporating any additional fine-tuning components. As it requires no extra GPU memory, it is particularly advantageous for practical deployment scenarios with limited computational resources. 

\vspace{-1mm}

\section{Experiments}
Our method is extensively evaluated on two food datasets, including VFN\cite{mao2020, he2023long} and UEC-256\cite{kawano14c} using YOLOv11 object detection rate and FID score as evaluation metrics. We show both quantitative and qualitative results including ablation studies. From different experiments, both modules in our method perform well to address the challenges in generating food images with multi-noun categories. Additionally, we show that our method can also generalize to non-food domains.

\subsection{Datasets}
We conduct our experiments on VFN \cite{mao2020, he2023long} and UEC-256 \cite{kawano14c} dataset because they have bounding boxes for each food that can be used to train a detection model to detect the objects in the generated images as an evaluation metric. 

\noindent\textbf{VFN\cite{mao2020, he2023long}:} The VFN dataset consists of 186 food categories with approximately 40k images. Among all food categories, there are 37 food categories considered as multi-noun. Each image is annotated with food category names and corresponding bounding box locations. The food categories represent commonly consumed foods selected from the What We Eat In America (WWEIA) database \cite{eicher2017}. We split the dataset into 90\% for fine-tuning and 10\% for testing.

\noindent \textbf{UEC-256 \cite{kawano14c}:} The UEC-256 dataset contains 256 food categories in Japanese cuisine. The food images are annotated with bounding boxes and food types. There are 133 out of 256 food categories considered as multi-nouns. We randomly select 90\% of the images for fine-tuning and 10\% for testing, consistent with the VFN dataset.

\subsection{Evaluation Metrics}
To evaluate whether generated images contain the required food items and any other redundant generated food objects, we employ both object detection and distribution-based metrics. For the object-level evaluation, we use YOLOv11 \cite{yolo2023}, an object detection model fine-tuned on the training subset of food datasets. We report standard detection metrics including \textbf{precision}, \textbf{recall} and \textbf{F-1 score}. 
We also measure the overall image quality using the \textbf{Fréchet Inception Distance (FID)} \cite{heusel2017gans} score which compares the feature distribution of real and generated images. Together, these metrics provide a comprehensive evaluation of both object existence and visual fidelity of the generated images. 

\subsection{Experiment Setup}
We fine-tune the Stable diffusion model\cite{Rombach_2022_CVPR} on the pre-trained weight (\textit{CompVis/stable-diffusion-v1-4}) using a learning rate of 1e-4 on the UNet and 1e-5 on the CLIP text encoder, with a batch size of 8 and 30 epochs. A weight decay of 1e-3 is applied to both components. 
A total number of 50 inference steps is used. 

As discussed in Sections~\ref{sec:rel_attr_bind} and~\ref{sec:FDKF}, attribute binding and text-encoder fine-tuning are closely related to our work, so we compare our method with the following related works: \textbf{Structured Diffusion}~\cite{feng2023trainingfree}, which addresses attribute binding via structured guidance at inference; \textbf{Syngen}~\cite{rassin2024linguistic}, an inference-time method that optimizes attention distributions to bind attributes to the correct objects; the latest \textbf{Stable Diffusion~3 (SD3)}~\cite{esser2024scaling}, which has more powerful text encoders (CLIP~\cite{radford2021learning} and T5~\cite{2020t5}) with a diffusion transformer backbone~\cite{peebles2023scalable}; \textbf{SD3 + Prompt Rewriting}~\cite{amir2023}, which rephrases compound food categories into natural descriptive sentences; and \textbf{TextCraftor}~\cite{li2024textcraftor}, which fine-tunes the CLIP text encoder for global image–text alignment.
For a fair comparison, we follow common practice to first fine-tune the Stable diffusion model with the text encoder frozen but with UNet activated. Next, we load the resulting weights into \textbf{Structured Diffusion} and \textbf{Syngen} for evaluation. 
We also conduct ablation studies for the proposed \textbf{FDALA} and \textbf{CFIG} on both datasets. 

\begin{table*}[t]
\begin{center}
\footnotesize
\caption{Comparison with related works for \textbf{multi-noun categories} in VFN and UEC-256 datasets} 
\begin{tabular}{c|c|cccc|cccc}
  \toprule
  & \multicolumn{4}{c}{VFN dataset} & \multicolumn{4}{c}{UEC-256 dataset} \\
  \cmidrule(lr){3-6} \cmidrule(lr){7-10}
  \makecell{Method} & \makecell{Text encoder \\ fine-tuning ?} & Precision \(\uparrow\) & Recall\(\uparrow\) & F-1 score\(\uparrow\) & \makecell{FID \\ score\(\downarrow\)} & Precision\(\uparrow\) & Recall\(\uparrow\) & F-1 score\(\uparrow\) & \makecell{FID \\ score\(\downarrow\)} \\
  \midrule
  \makecell{Real images (Reference Value)} & -- & 0.347 & 0.312 & 0.329 & -- & 0.776 & 0.395 & 0.524 & -- \\
    \hdashline 
  \makecell{Stable diffusion \cite{hu2022lora}} &\xmark& 0.248 & 0.195 & 0.218 & 86.2 & 0.232 & 0.079 & 0.118 & 36.3 \\
  \makecell{Structured diffusion \cite{feng2023trainingfree}} &\xmark& 0.296 & 0.243 & 0.267 & 84.8 & 0.198 & 0.072 & 0.106 & 39.0 \\
  Syngen \cite{rassin2024linguistic} &\xmark& 0.196 & 0.124 & 0.152 & 92.0 & 0.075 & 0.03 & 0.043 & 56.7 \\
  \makecell{{Stable Diffusion 3 (SD3) \cite{esser2024scaling}}} &\xmark& 0.344 & 0.311 & 0.327 & 83.9 & 0.223 & 0.192 & 0.206 & 45.1 \\
  \makecell{TextCraftor\cite{li2024textcraftor}} &\cmark& 0.375 & 0.329 & 0.350 & \textbf{72.7} & 0.49 & 0.201 & 0.285 & \textbf{35.5} \\
  \bottomrule
  \makecell{FoCULR(Ours)} &\cmark& \makecell{\textbf{0.457}} & \makecell{\textbf{0.433}} & \makecell{\textbf{0.445}} & \makecell{73.7} & \makecell{\textbf{0.533}} & \makecell{\textbf{0.217}} & \makecell{\textbf{0.308}} & \makecell{\textbf{35.5}}\\
  \bottomrule
\end{tabular}
\label{tab:yolo_detect_multi_noun}
\end{center}
\vspace{-5mm}
\end{table*}

\begin{table}[t]
\begin{center}
\footnotesize
\caption{Ablation studies of our method for VFN dataset on \textbf{multi-noun categories:} Our method results are shown with average value from 3 runs $\pm$ standard deviation} \label{tab:ablation_vfn_multi-noun}
\begin{tabular}{cccccc}
\hline
\makecell{FDALA} & \makecell{CFIG} & \makecell{Text \\ encoder \\ fine-tuning ?} & \makecell{Precision\(\uparrow\)} & \makecell{Recall\(\uparrow\)} & \makecell{F-1 \\ score\(\uparrow\)} \\
\hline
\xmark &\xmark&\xmark& \makecell{0.248} & \makecell{0.195} & \makecell{0.218} \\
\xmark&\cmark&\xmark& \makecell{0.31\\ $\pm$ 0.013} & \makecell{0.253\\ $\pm$ 0.014} & \makecell{0.279\\ $\pm$ 0.014} \\
\cmark&\xmark&\cmark& \makecell{0.451\\ $\pm$ 0.012} & \makecell{0.432\\ $\pm$ 0.017} & \makecell{0.441\\ $\pm$ 0.015} \\
\cmark&\cmark&\cmark& \makecell{\textbf{0.457}\\ $\pm$ 0.014} & \makecell{\textbf{0.433}\\ $\pm$ 0.016} & \makecell{\textbf{0.445}\\ $\pm$ 0.015} \\
\hline
\end{tabular}
\end{center}
\vspace{-5mm}
\end{table}

\begin{table}[t]
\begin{center}
\footnotesize
\caption{Ablation studies of our method for UEC-256 dataset on \textbf{multi-noun categories:} Our method results are shown with average value from 3 runs $\pm$ standard deviation} \label{tab:ablation_uec_multi-noun}
\begin{tabular}{cccccc}
\hline
\makecell{FDALA} & \makecell{CFIG} & \makecell{Text \\ encoder \\ fine-tuning ?} & \makecell{Precision\(\uparrow\)} & \makecell{Recall\(\uparrow\)} & \makecell{F-1 \\ score\(\uparrow\)} \\
\hline
\xmark&\xmark&\xmark& \makecell{0.232} & \makecell{0.079} & \makecell{0.118} \\
\xmark&\cmark&\xmark& \makecell{0.239\\ $\pm$ 0.004} & \makecell{0.09\\ $\pm$ 0.007} & \makecell{0.131\\ $\pm$ 0.006} \\
\cmark&\xmark&\cmark& \makecell{0.511\\ $\pm$ 0.016} & \makecell{0.215\\ $\pm$ 0.008} & \makecell{0.303\\ $\pm$ 0.007} \\
\cmark&\cmark&\cmark& \makecell{\textbf{0.533}\\ $\pm$ 0.018} & \makecell{\textbf{0.217}\\ $\pm$ 0.001} & \makecell{\textbf{0.308}\\ $\pm$ 0.006} \\
\hline
\end{tabular}
\end{center}
\vspace{-5mm}
\end{table}

\begin{figure*}[t]
    \centering\includegraphics[width=1\linewidth]{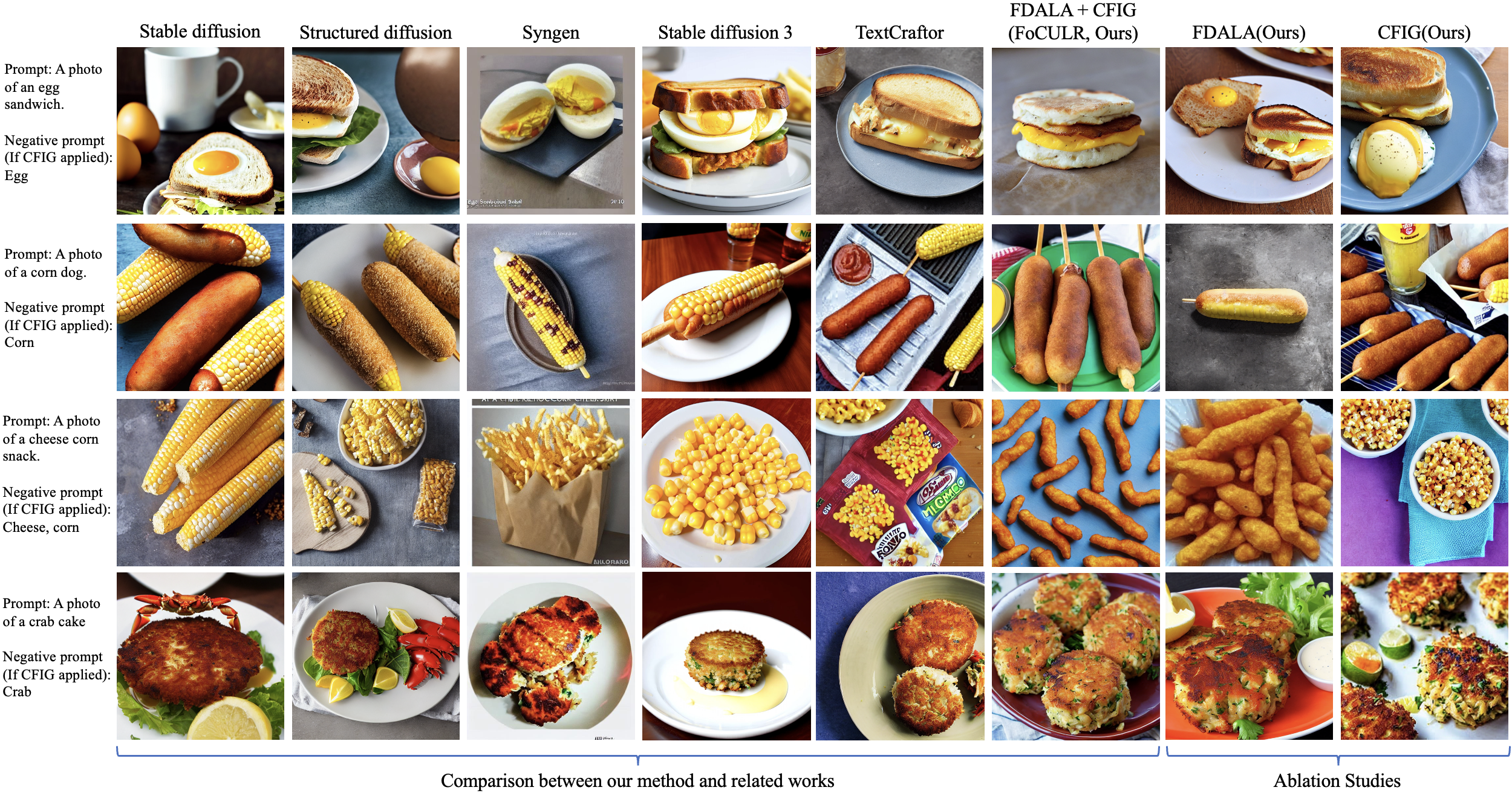}
    \caption{Comparison of food image generation results to related works (Stable diffusion, Structured diffusion, Syngen, TextCraftor and Stable Diffusion 3) and ablation studies on our methods (CFIG, FDALA): Our method performs well by generating food objects with multi-noun categories without redundant food objects while the prior works tend to generate irrelevant food objects.}
    \label{fig:comparison}
    \vspace{-3mm}
\end{figure*}

\begin{figure}[htp]
    \centering
    \includegraphics[width=0.95\linewidth]{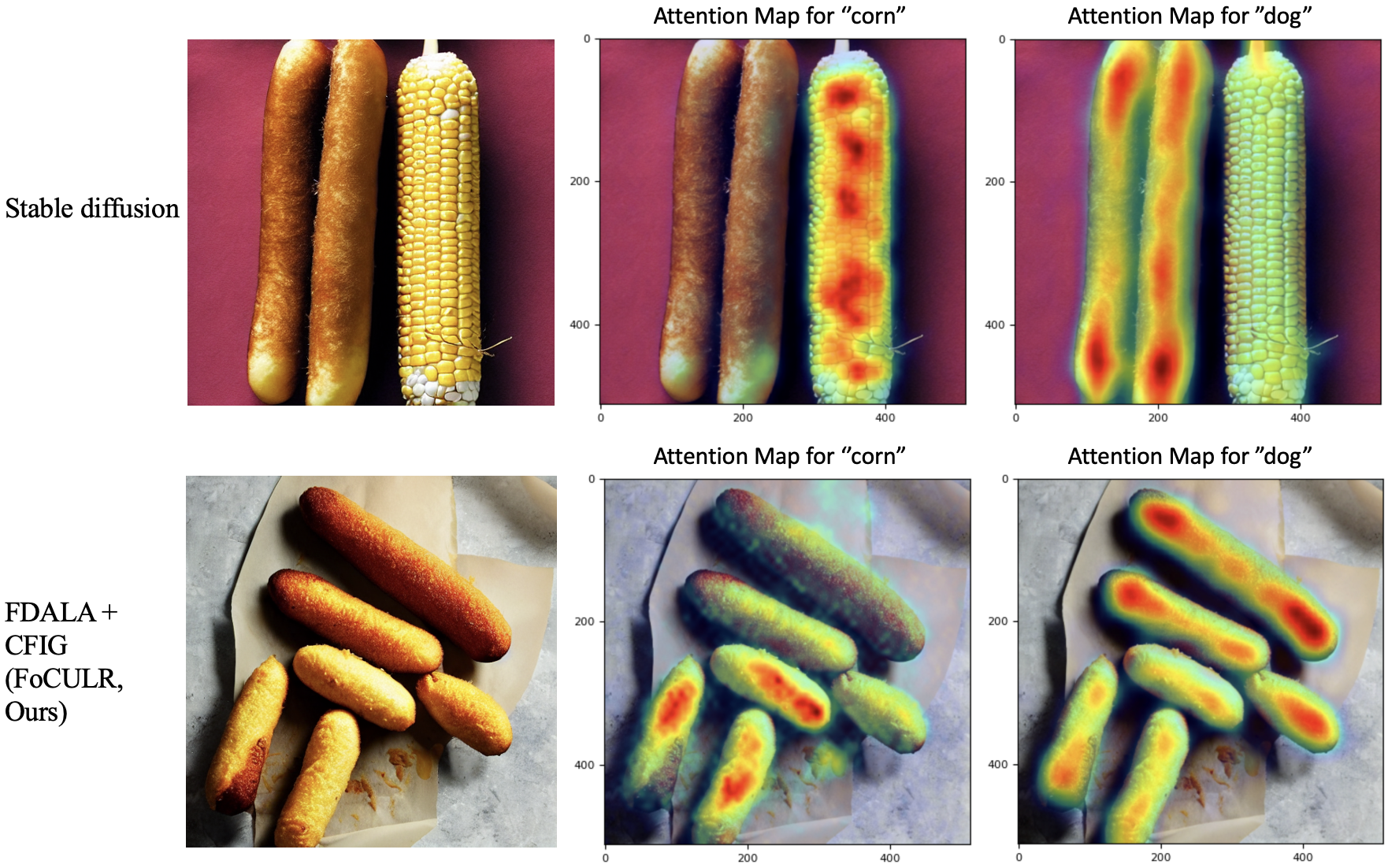}
    \caption{Attention map for generated image with prompt of ``A photo of a corn dog." and negative prompt of ``corn" if CFIG applied.}
    \label{fig:attention_corn_dog}
    \vspace{-3mm}
\end{figure}

\begin{figure}[htp]
    \centering
    \includegraphics[width=0.95\linewidth]{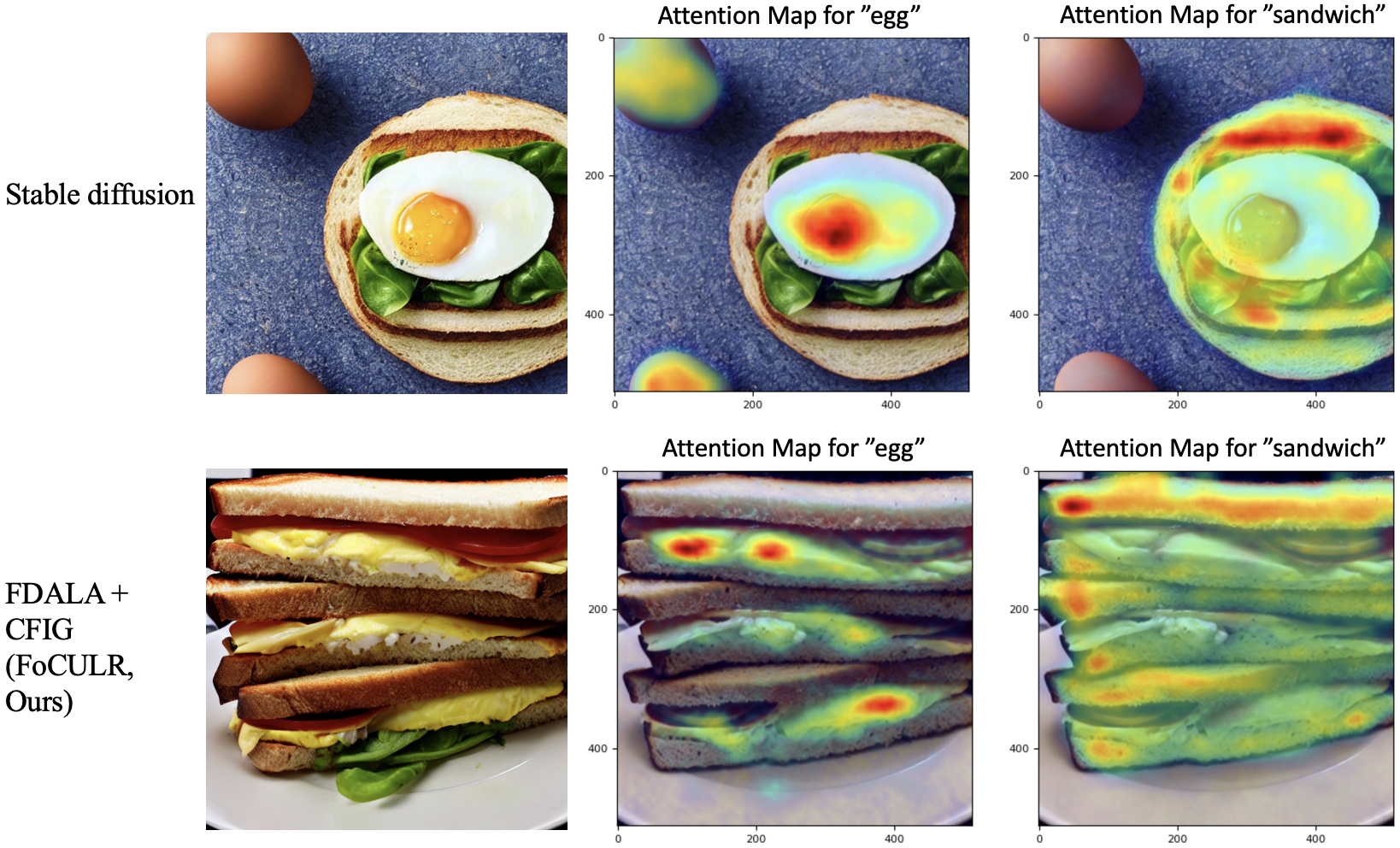}
    \caption{Attention map for generated image with prompt of ``A photo of a egg sandwich." and negative prompt of ``egg" if CFIG applied.}
    \label{fig:attention_egg_sandwich}
    \vspace{-3mm}
\end{figure}

\begin{figure}[htp]
    \centering
    \includegraphics[width=0.95\linewidth]{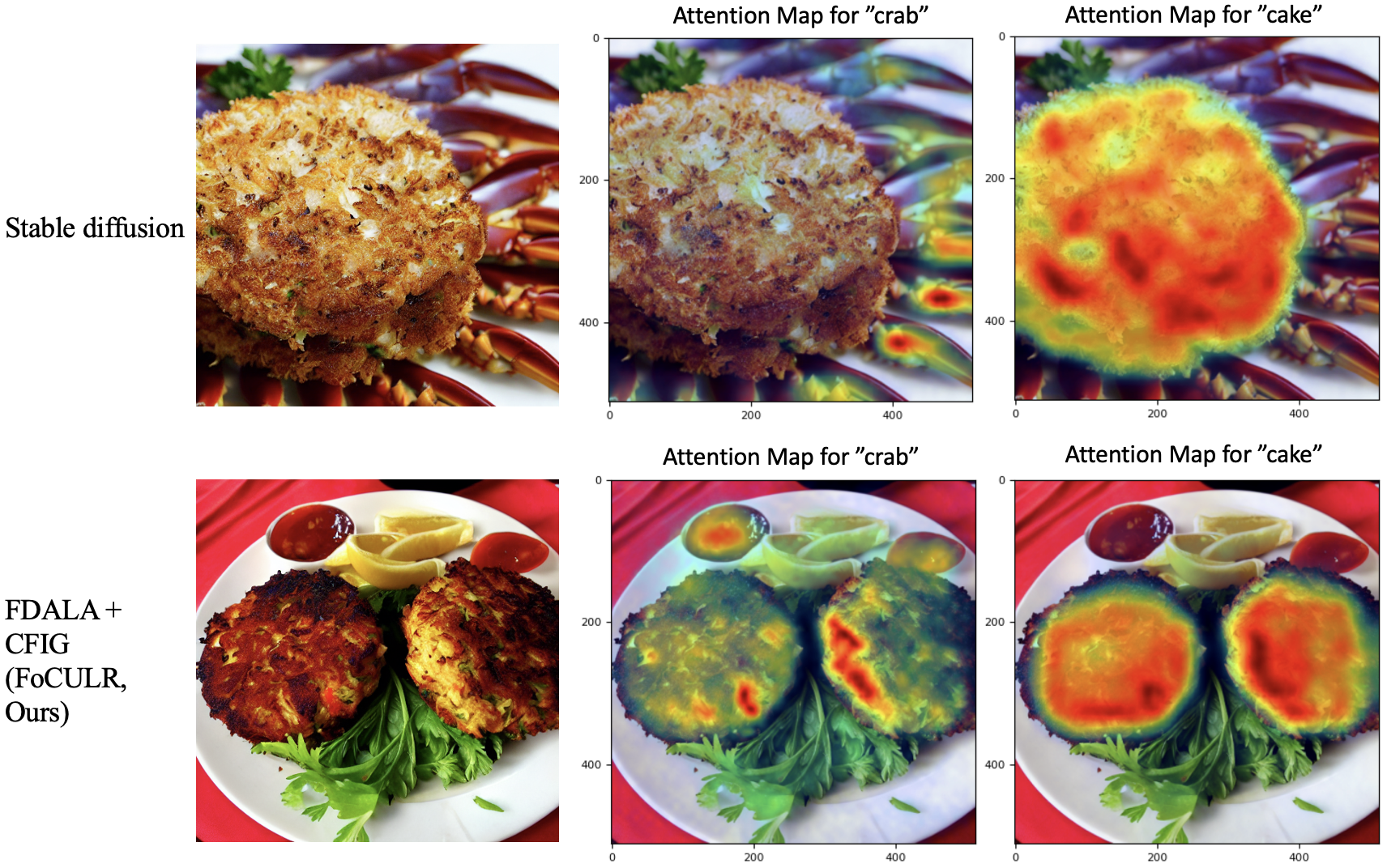}
    \caption{Attention map for generated image with prompt of ``A photo of a crab cake." and negative prompt of ``crab" if CFIG applied.}
    \label{fig:attention_crab_cake}
    \vspace{-3mm}
\end{figure}

\begin{figure}[htp]
    \centering
    \includegraphics[width=0.95\linewidth]{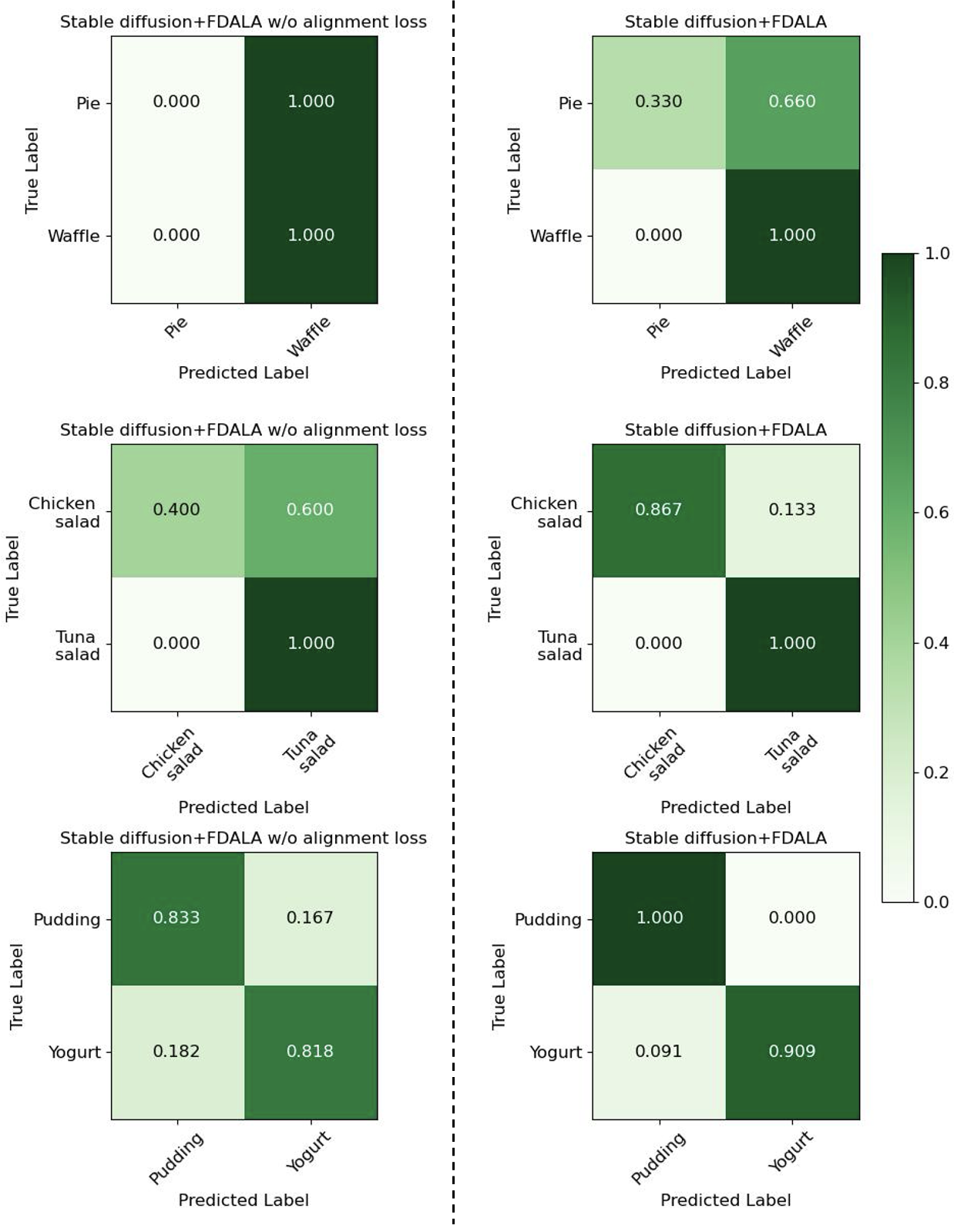}
    \caption{Confusion matrix with classification accuracy by YOLOv11 model between easily confused categories with/without image-concept alignment: darker color on the diagonal indicates better result.}
    \label{fig:confus_mat}
    \vspace{-3mm}
\end{figure}

\subsection{Quantitative Results}
\label{sec:exp_results}

\noindent\textbf{Overall Quantitative Results:} Table \ref{tab:yolo_detect_multi_noun} presents quantitative results for multi-noun categories. When generating images from multi-noun food categories, our proposed method, FoCULR, significantly outperforms the baseline (e.g., precision of 0.457 for FoCULR vs. 0.375 for TextCraftor on VFN), highlighting its capability to produce coherent and unified representations of food items with multi-noun. This improvement primarily comes from FoCULR’s enhanced ability to encode food-domain knowledge and assign semantically appropriate attention weights to each noun within compound phrases, resulting in fewer redundant objects and more accurate layouts during denoising. These results confirm the effectiveness of our combined approach—fine-tuning the text encoder along with targeted negative-prompt scheduling—in reducing redundant object generation. The slight FID increase likely stems from reduced background diversity due to our focus on object correctness (with early negative prompts).

Additional results covering all categories can be found in the supplementary material Section B, which shows that our proposed method improves the overall image generation performance among all categories.


\noindent\textbf{Ablation Studies:} Tables \ref{tab:ablation_vfn_multi-noun} and \ref{tab:ablation_uec_multi-noun} assess the individual contributions of our two modules. CFIG alone can outperform the baseline model without text encoder finetuning, which also does not require additional computational cost. FDALA alone can significantly improve beyond the baseline model. Finally, FDALA+CFIG achieves the best results, illustrating that the two modules complement each other. Qualitative results in Figure \ref{fig:comparison} can further verify the effectiveness of different modules in our proposed method, as detailed in Section \ref{sec:qual}.

\subsection{Qualitative Results}
\label{sec:qual}
\noindent\textbf{Generated Image Results:} Figure \ref{fig:comparison} shows side-by-side image generation results for four multi-noun food categories: ``egg sandwich," ``corn dog,” ``cheese corn snack," and ``crab cake.” We compare five baselines (Stable Diffusion, Structured Diffusion, Syngen, TextCraftor and Stable Diffusion 3) against individual components of our proposed FoCULR method (FDALA only, CFIG only, and FDALA+CFIG).

Images generated by baseline methods contain redundant or extraneous food items. For example, Stable Diffusion, Structured Diffusion, and Syngen generate separate eggs for egg sandwiches, extra corn around corn dogs, and additional crab components for crab cakes. Despite SD3’s stronger text encoders, it still produces redundant surface details (e.g., corn kernels on a corn dog) even when the object is generated as a single entity. Although TextCraftor partially reduces these redundant objects, unnecessary details persist, such as scattered corn kernels alongside a coherent corn dog or unobvious crab leg on the right. This issue arises because TextCraftor prioritizes global text-to-image relationships, neglecting local visual consistency.

In contrast, our method improves image coherence through complementary strategies. More specifically, FDALA alone significantly reduces redundant objects but may still produce minor irrelevant surface details, such as corn kernels on corn dog. However, it is still better than TextCraftor because it can eliminate more unnecessary food elements, i.e., no crab leg beside the crab cake, no additional corn beside the corn dog. CFIG alone, without fine-tuning the text encoder, explicitly suppresses redundant object layouts but still generates less accurate object appearances, such as popcorn-like cheese corn snacks. Combining FDALA and CFIG produces best results, consistently eliminating redundant objects and clearly depicting unified food items, highlighting the complementary effectiveness of our integrated approach.

\subsection{Additional Analysis}
\label{sec:qual2}
\noindent\textbf{Attention Map Visualization:}  
To understand the advantage of our proposed method, we visualize per‑token attention maps on generated images to show FoCULR can reflect contextual weight for a food in Figures~\ref{fig:attention_corn_dog}, \ref{fig:attention_egg_sandwich}, and \ref{fig:attention_crab_cake}.
Under the baseline Stable Diffusion model, each noun in a compound prompt activates a separate region. 
In contrast, our method encourages shared attention across related tokens, enabling the model to treat compound food labels as coherent objects rather than independent parts.



\noindent\textbf{Inter-Class Similarity Issue:} Figure \ref{fig:confus_mat} shows confusion matrices with classification accuracy by YOLOv11 model for three commonly mixed-up food pairs—“pie” vs. “waffle,” “chicken salad” vs. “tuna salad,” and “pudding” vs. “yogurt” with/without image-concept alignment loss in FDALA, as introduced in Section \ref{sec:FDKF}. 
Without the alignment loss, the model almost always confuses each pair, 
while incorporating the patch-level alignment loss dramatically boosts correct detections. 
These results confirm the need of image-concept alignment loss in our FDALA module. Although our primary focus is on multi-noun categories, addressing inter-class similarity issue remains important for maintaining overall image fidelity and semantic consistency.



\begin{figure}[htp]
    \centering
    \includegraphics[width=0.95\linewidth]{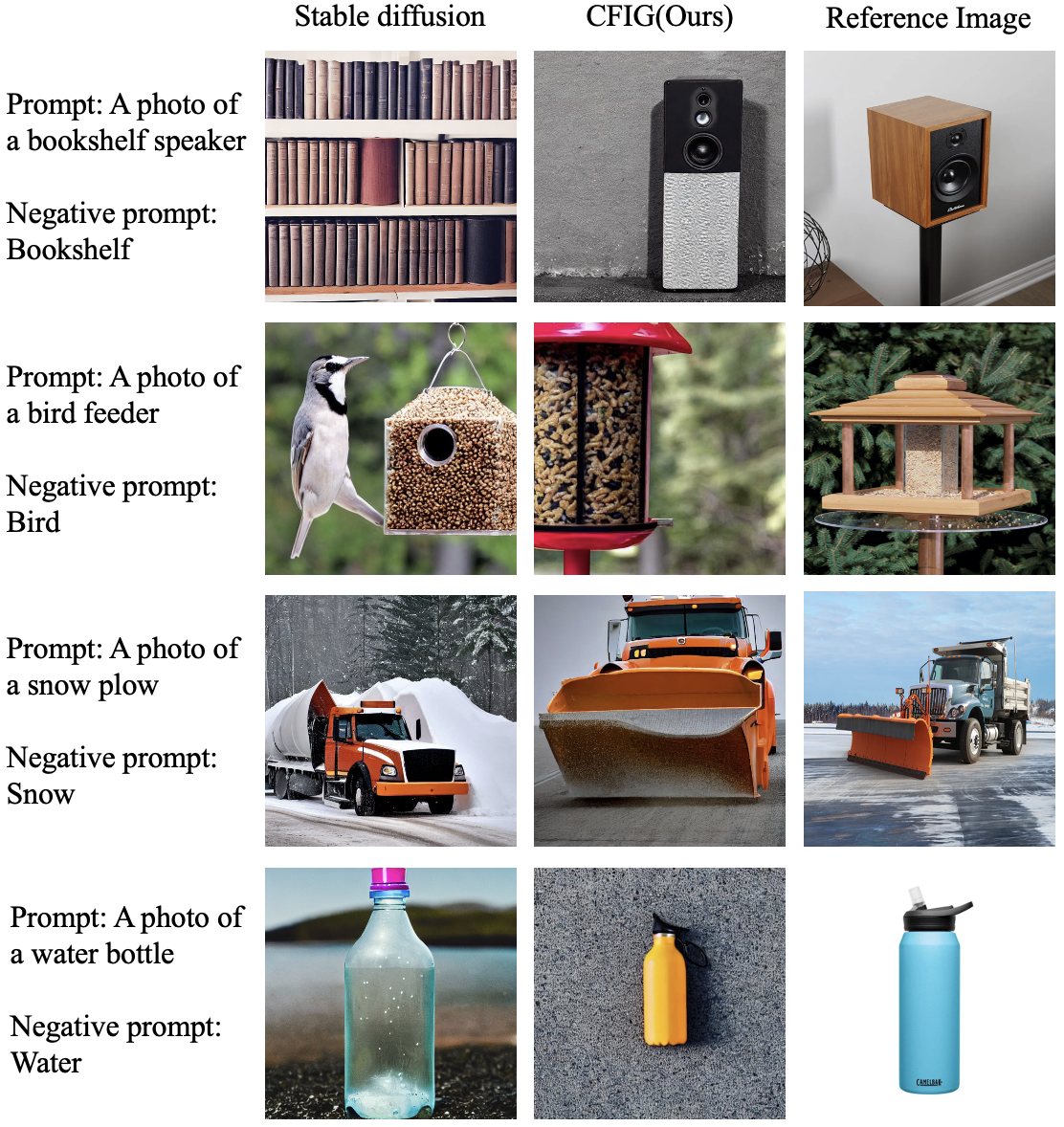}
    \caption{Example generated images in non-food domain}
    \label{fig:general}
    \vspace{-3mm}
\end{figure}

\subsection{Generated Image Results in Non-Food Domains}

To demonstrate the generalizability of our CFIG approach beyond food domain, Figure \ref{fig:general} shows four non-food, multi-noun prompts: “bookshelf speaker,” “bird feeder,” “snow plow,” and “water bottle.” In each case, the vanilla Stable Diffusion model misinterprets the compound prompt, producing disjoint visual elements. For example, “bookshelf speaker” results in a bookshelf filled with books rather than a speaker designed for shelf placement, while “snow plow” yields drifting snow beside a plow blade rather than a unified snow-plow vehicle.

By applying CFIG’s head-noun scheduling (i.e., negative-prompting “bookshelf,” “bird,” “snow,” or “water” during the initial denoising steps), our method forces the model to first anchor the core object layout (speaker, feeder, plow, bottle) before reintroducing modifiers. As a result, CFIG yields coherent single-object outputs that closely match the reference images. Although the multi-noun problem is less pronounced in these non-food domains, these examples affirm the broader applicability of our head-driven negative-prompt strategy in any compound noun prompt for preventing disjointed generations and enforcing unified object layouts.

\begin{figure}[htp]
    \centering
    \includegraphics[width=0.95\linewidth]{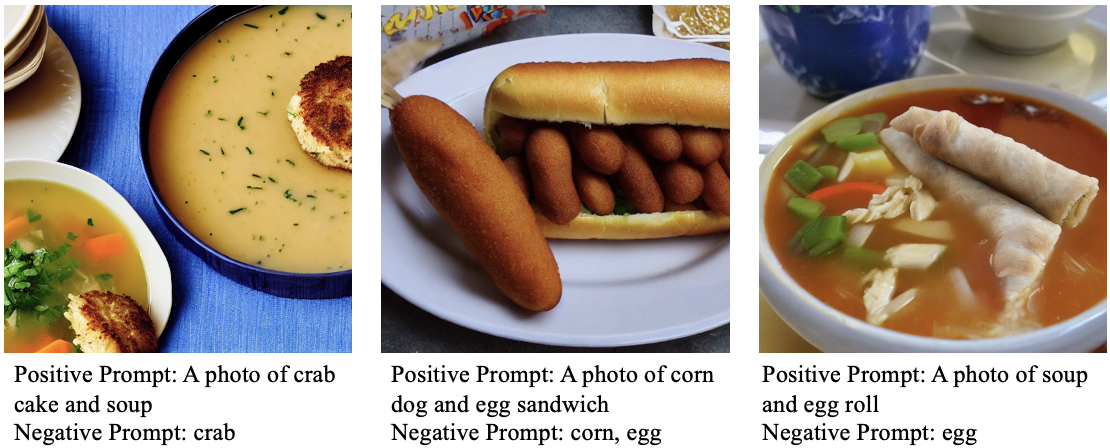}
    \caption{Failed Examples of generated images using our method}
    \label{fig:limit}
    \vspace{-3mm}
\end{figure}

\subsection{Limitations}
Figure \ref{fig:limit} shows the failed examples of generated images using our method. When coming to generate multiple foods with multi-noun categories in an image, our method could cause object entanglement issue (e.g. crab cake and soup), making two objects fused together. This will be served as the future work in this paper.

\vspace{-1mm}

\section{Conclusion}

In this paper, we tackled the multi-noun category problem in food image generation by proposing the Food Domain Adaptation via Local Alignment (FDALA) and Core-Focused Image Generation (CFIG) modules, which effectively mitigate redundant object generation. Future work will explore multi-noun categories in compositional food image generation, aiming to enhance the spatial relationships between multiple food items in generated images.


{\small
\bibliographystyle{ieee_fullname}
\bibliography{egbib}
}

\end{document}